# BRICE: Brain-inspired Recurrent Illusory Contour Encoder for Point Cloud Completion

**Yunlei Wang [1,†], Dawei Li [1,†], and Xue-song Tang [1,*]**

1 School of Information and Intelligent Science, Donghua University, Shanghai, China 201620

† Contributed equally; * Corresponding author

## Abstract

Mainstream point cloud completion methods are unanimously based on deep learning architectures. On one hand, these methods tend to repetitively stacking attention modules to enhance network performance, which drifts away from the neat design of unified and interpretable networks; on the other hand, nearly all existing architectures rely solely on feedforward connections to propagate information, neglecting feedback from deeper layers. The recurrent neural architecture, formed by feedforward-feedback loops, have shown its ability to generate Illusory Contours (ICs) in the V1 brain region of mammals; and ICs are essentially a completion mechanism of the brain for visual information. Motivated by this, we design a Brain-inspired Recurrent Illusory Contour Encoder (BRICE) to "imagine" what is missing in the incomplete 3D point clouds. Building upon the foundation model of Predictive Coding (PC), the plug-and-play nature enables BRICE to be seamlessly integrated into mainstream point cloud completion baselines. Experimental results show that BRICE can enhance various completion baselines on different 3D datasets such as MVP2K, PCN, and PlantPCom. Our code: https://github.com/wangyunlei-wyl/BRICE.

## Introduction

The point cloud is a general format to represent 3D data. Point clouds acquired from real-world scenarios often exhibit sparse or missing structures due to occlusion, sensor resolution constraints, and operations such as down-sampling. The incompleteness severely affects data fidelity and applicability. The task of point cloud completion typically refers to recovery of a complete structure from an incomplete and sparse point cloud. The recovered point cloud offers better data to subsequent 3D vision tasks such as point cloud classification (Uy et al. 2019), segmentation (Kolodiazhnyi et al. 2024), and 3D object detection (Li and Chen 2025) with potentially promising results.

Mainstream point cloud completion methods are unanimously deep-learning-based, and can be broadly classified into several types. The first type adopts the standard encoder-decoder architecture assisted with attention modules/Transformers. As a representative work among those, PCN (Yuan et al. 2018) presented a coarse-to-fine, two-stage generation strategy that operates directly on raw point clouds.

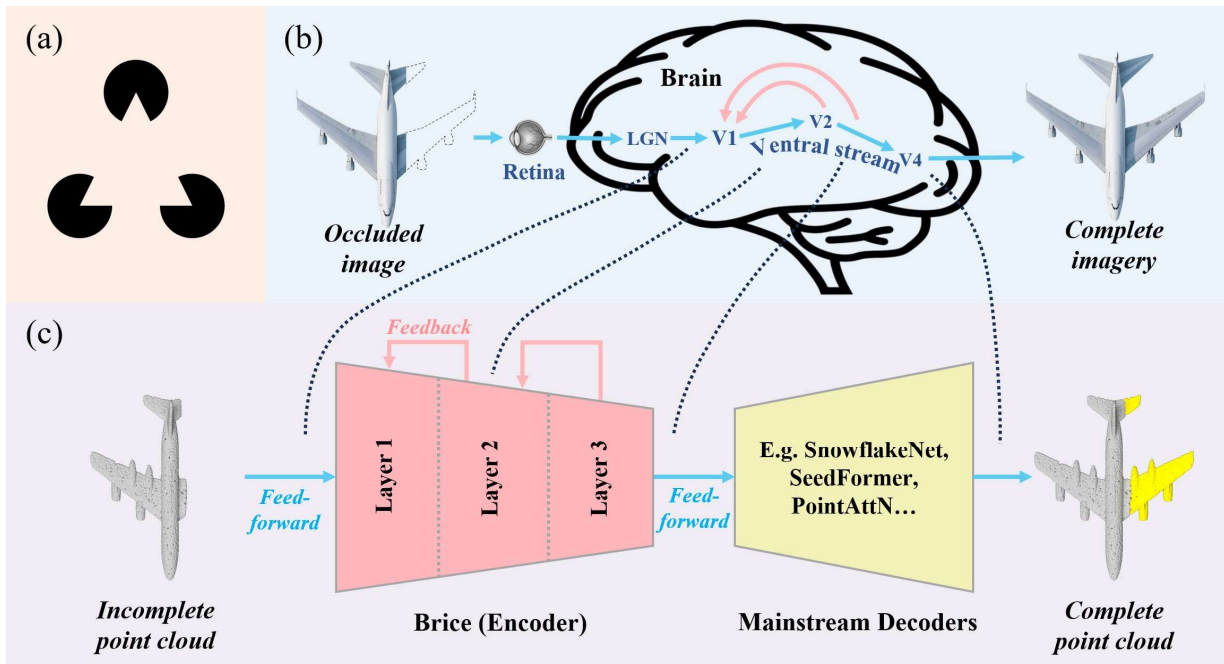

Figure 1: Illusory Contour and the design of our BRICE. (a) The Kanizsa triangle; (b) A possible brain mechanism to recognize missing 2D visual information; (c) A diagram of our plug-and-play BRICE for point cloud completion.

Subsequent variants of this type (Tchapmi et al. 2019; Xie et al. 2020; Huang et al. 2020; Yu et al. 2021; Xiang et al. 2021) improved their ability to capture global shape context and enhanced the understanding of rich 3D geometric details by introducing stacked Transformer-like modules. The second category utilizes local point-wise relationships of inter-layer and intra-layer (Wen et al. 2021, 2022; Zhou et al. 2022; Yan et al. 2022; Rong et al. 2024) to enhance completion performance. Those approaches effectively facilitate information transfer and feature interaction with carefully designed cross-resolution interaction mechanisms or feature aggregation modules, enabling the networks to infer the missing data. The methods of the third type are either heuristic (Yan et al. 2025; Kim and Cho 2026; Li et al. 2026) or tailored under specific theories (Xi et al. 2026). By introducing the real-world properties or physical mechanisms (such as symmetry priors and diffusion dynamics) into modular design, those methods reason about the geometric structures of missing regions with enhanced interpretability, and ultimately generating point clouds with high fidelity and plausible topologies.

Although new methods for point cloud completion continue to spring up, several challenges remain unresolved. First, the architecture design tends to stacking complex deep attention modules to elevate the learning performance, particularly at decoding. The stacking relies heavily on empirical trial and error, which drifts away from the neat design of unified and interpretable networks. Second, despite several recent heuristic solutions (Zhou et al. 2025; Kim and Cho 2026; Yu et al. 2025), mainstream studies still lack rigorous theoretical guidance, leaving the intrinsics of 3D completion largely under-explored. Third, the bulk of existing studies employ only feedforward connections for information propagation, which overlooks the feedback as top-down supervision from deeper layers. The recurrent framework driven by feedforward-feedback loops has been neurobiologically proven to mediate visual information completion (e.g., the Illusory Contours) inside the V1 visual cortex of mammal brains. This encourages us to think whether it is possible to step out of the conventional layer-stacking architecture and switch to a new module that is effective and explainable?

Illusory Contours (ICs) refer to the situation that the brain interprets shapes that do not actually exist in the physical input during perception of edges in visual signals. Recent studies unveiled that a subset of neurons in the V1 primary visual cortex of mammal brains can exhibit emergent responses to illusory stripes in images (e.g., humans can recognize the triangle inside Kanizsa pattern in **Fig. 1(a)**), rather than merely to local image textures. The neurons for pattern completion in primary cortical regions may selectively enhance certain image signals with predictive priors (Shin et al. 2025). This gives rise to the "Gestalt" perception, i.e., the whole is something else than the sum of its parts. Furthermore, some specialized structures in V1 L2/3 are not only characterized by standard feedforward connections but are also governed by feedback connections from high-level brain regions, forming a recurrent connectivity (as shown in **Fig. 1(b)**). Researchers in computer vision (Pang et al. 2021) also confirms the existing of similar response to ICs when processing images. The conventional CNNs using purely feedforward connections in image classification fail to perceive ICs in images, whereas adding top-down feedback generates emergent IC inferences. We guess that since the brain or V1-inspired artificial neural networks form triangular perception when processing Kanizsa patterns in 2D images, they also might be able to imagine the missing part in another type of visual data—3D point cloud. We draw inspiration from brain for "guessing" the incomplete/missing part on point clouds under a recurrent architecture, and propose BRICE—Brain-inspired Recurrent Illusory Contour Encoder. The main contributions are as follows:

- Since V1 is the shallow layers in visual perception, BRICE should correspond to the encoder in the network. Specifically, BRICE takes the hierarchical shape of Predictive Coding (PC) model of the primary visual cortex. It incorporates both standard feedforward path and feedback guidance from deeper layers, forming a recurrent architecture that serves completion (as shown in **Fig. 1(c)**).
- BRICE is designed to be plug-and-play. It can be seamlessly integrated into existing mainstream point cloud completion networks by replacing their original encoders for enhancing performance. On 3D datasets covering both man-made objects and natural crops, BRICE can improve the performances of various mainstream baselines for point cloud completion.
- Transferring pretrained BRICE (with classification task) to a new point cloud completion task can further improve the completion performances of multiple baselines. This suggests that learning the statistical regularities of 3D real-world shapes beforehand helps the BRICE encoder to make reasonable predictive priors about missing parts.

## Related Work

### Point Cloud Completion

The 1st-gen point cloud completion networks have standard encoder-decoder structure. For instance, PCN (Yuan et al. 2018) and TopNet (Tchapmi et al. 2019) established a coarse-to-fine, two-stage framework. However, their feature extraction relied primarily on global max-pooling, leading to the loss of fine-grained features. Subsequent studies (Yu et al. 2021; Xiang et al. 2021; Chen et al. 2023b; Yu et al. 2024; Zhong et al. 2025; Fang et al. 2025; Wei et al. 2025) incorporated attention mechanisms to improve feature extraction. For example, PointTr (Yu et al. 2021) pioneered the conversion of point clouds into feature sequences and utilized a Transformer decoder for enhancing the capacity to learn shape structures. SnowflakeNet (Xiang et al. 2021) proposed a Skip-Transformer to capture global shape contextual information and geometric spatial relationships.

To explicitly model the spatial continuity and structural consistency across multi-scale features, several methods (Wen et al. 2021, 2022; Zhou et al. 2022; Yan et al. 2022; Rong et al. 2024) established local relationships from hierarchical interactions to enhance point cloud completion performance. Among those, SeedFormer (Zhou et al. 2022) proposed Patch Seeds that not only captures global topology from local inputs across layers but also preserves detailed local information for recovering. CRA-PCN (Rong et al. 2024) proposed a cross-resolution Transformer that enables intra- and inter-layer cross-resolution feature interactions.

To enhance the interpretability, recent works have focused on heuristics (Yan et al. 2025; Kim and Cho 2026; Li et al. 2026) or specific theoretical frameworks (Xi et al. 2026). Symmetry is a common feature on man-made objects. SymmCompletion (Yan et al. 2025) introduced point-wise local symmetric transformations to fill missing regions with intact key geometries from the input. Inspired by NeRF, DANCE (Kim and Cho 2026) generated candidate points based on multi-view ray sampling and employed a Transformer decoder to infer the geometric features of missing regions. 3DDM (Xi et al. 2026) proposed a physically-based anisotropic 3D diffusion model aiming to conduct point cloud completion with fewer diffusion steps. However, current models require heavy computation of attention matrices across deep layers; and neglect the potential of direct completion by leveraging prior knowledge from recurrent loops with feedback supervision from deep layers.

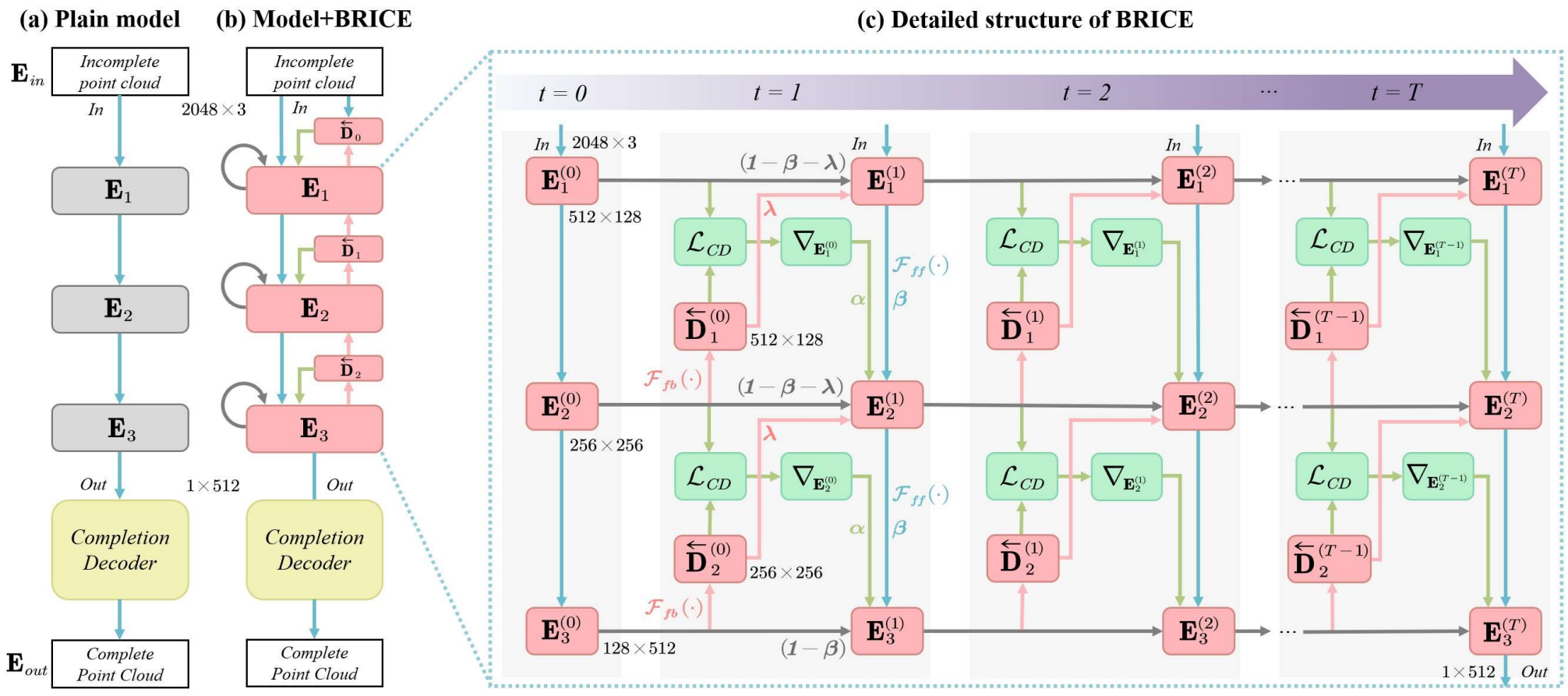

Figure 2: The detailed structure of BRICE. (a) shows a typical feedforward point cloud completion model; (b) shows the same completion model using BRICE as the encoder; (c) provides a detailed visualization of BRICE's internal calculations.

### Brain's Response to Illusory Contours

Neuroscience studies (Pak et al. 2020; Pang et al. 2021; Shin et al. 2025) has discovered the brains of different mammals responded emergently to ICs. These responses are characterized by occurring directly in low-level brain regions while depending on the feedback from higher-level regions to form a recurrent structure. These observations have inspired research in computer vision. Several studies (Pang et al. 2021; Choksi et al. 2021; Wen et al. 2018) incorporated similar recurrent responses to ICs into deep networks to reconstruct images. For example, a brain-inspired Predictive Coding (PC) model (Pang et al. 2021) used dynamic recurrent network to complete missing contours in simple images. Building on this work, PREDIFY (Choksi et al. 2021) applied the PC-like model to the VGG deep network, and proved the visual completion mechanism could enhance image reconstruction and denoising. Similarly, DPCN (Wen et al. 2018) also adopted the feedforward-feedback neural architecture to carry out image classification. On the CIFAR10/100 datasets, DPCN achieved comparable performance with only 7 layers against several ultra-deep networks. We are curious to learn whether V1-inspired architecture can be generalized from completion of 2D images to 3D data. For the first time, we design a plug-and-play recurrent encoder to improve existing 3D completion networks.

## Method

### Architecture of BRICE

The architecture of our BRICE is shown in **Fig. 2**. It is essentially a plug-and-play module that can replace the encoder component of various mainstream point cloud completion networks to further enhance performance. Inspired from how the recurrent V1 connectivity of brain response to ICs, BRICE has a three-layer encoding unit modeled under Predictive Coding. Each layer features not only feedforward pathway but also feedback from the deeper layer, forming a recurrent structure. The output of BRICE is directly fed into decoders of various point cloud completion networks, and improves the performances of original networks.

Unlike the encoders in plain models (**Fig. 2(a)**), the most distinctive trait of BRICE is its recurrent computation. As shown in **Fig. 2(b)**, each layer is connected to the previous layer via a sub-decoding feedback module $\overleftarrow{\mathbf{D}}$. Each layer passes feedforward information and meanwhile receives feedback guidance from higher-level regions, enabling each layer to predict/reconstruct the original input to itself. Due to inter-layer feedbacks, the computational flow must be carried out through a carefully orchestrated series of iterations to ensure orderly update. Therefore, we resort to the PC model for iterative computation, where the count of iterations is denoted by the recurrent time $T$. We use **Fig. 2 (c)** to illustrate the detailed computation pipeline from $\mathbf{E}_1$ to $\mathbf{E}_3$.

Furthermore, we can also pre-train BRICE on other tasks such as point cloud classification, and then transplant it to the completion task. Initializing BRICE with shared knowledge further enhances the performance of completion.

## Predictive coding for updating

The recurrent computations of BRICE follow the PC principle (Rao and Ballard, 1999) derived from primary visual cortex. This aligns with the brain's mechanism when processing ICs. BRICE is a dynamic system. Without increasing the number of physical layers in the network, it continuously performs self-prediction along the temporal dimension and updates its internal feature representations in iterations. After computation, the features of deepest layer can ultimately be processed by decoders of mainstream completion networks to reconstruct the complete point cloud. In the subsequent formulation of BRICE, we define $\mathbf{E}_l^{(t)}$ as the features of the $l$ -th layer at recurrence time $t \in \{0, \ldots, \mathrm{T}\}$. As shown in **Fig. 2(c)**, for any intermediate layer $\mathbf{E}_l$, the update equation at time $t$ is as follows:

$$\mathbf{E}_l^{(t)} = \beta \overrightarrow{\mathbf{E}}_l^{(t)} + \lambda \overleftarrow{\mathbf{D}}_l^{(t-1)} + (1-\beta-\lambda)\mathbf{E}_l^{(t-1)} - \alpha \overleftarrow{\mathbf{G}}_l^{(t-1)}, \quad (1)$$

where the hyperparameters $\beta$, $\lambda$, and $\alpha$ represent the feedforward coefficient, the feedback correction coefficient, and the reconstruction error coefficient, respectively. It should be noted that when it reaches the deepest layer $\mathbf{E}_3$, the feedback should set as $\lambda = 0$ because there are no higher layers.

**The feedforward process.** For the feedforward pathway (blue arrows in **Fig. 2**), BRICE propagates the latest state from the shallow layer to the deeper layers. As shown in **Fig. 2(c)**, for layer $l$, the feedforward input comes from the output $\mathbf{E}_{l-1}^{(t)}$ of layer $l-1$ at the current time $t$. When $l=1$, the feedforward input is the incomplete point cloud. We use a feedforward feature mapping function $\mathcal{F}_{ff}(\cdot)$ for feature processing; it consists of a set abstraction operation similar to that of the PointNet++ (Qi et al. 2017) and a standard Transformer for feature extraction. The feedforward process can be formulated as follows:

$$\overrightarrow{\mathbf{E}}_l^{(t)} = \mathcal{F}_{ff}(\mathbf{E}_{l-1}^{(t)}). \quad (2)$$

The feedforward process is further modulated by hyperparameter $\beta$ and is integrated into the current layer by **Eq. (1)**, providing a feature fusion.

**The feedback process.** During the feedback process, the network leverages features from the next layer to predict the features of the current layer, and uses the gradient of the error between prediction and actual features to guide the parameter updates of the current layer. Specifically, the feature $\mathbf{E}_l^{(t)}$ from the $l$ -th layer is passed through an inter-layer feedback sub-decoder, denoted as $\mathcal{F}_{fb}(\cdot)$, to generate a prediction $\overleftarrow{\mathbf{D}}_{l-1}^{(t)}$ for the layer $l-1$ (as shown in **Eq. (3)**). This feedback decoder is a cascade of three components—the first and third components are Multi-layer Perceptrons (MLPs), while the second component is an up-sampling module implemented via deconvolution.

$$\overleftarrow{\mathbf{D}}_{l-1}^{(t)} = \mathcal{F}_{fb}(\mathbf{E}_l^{(t)}). \quad (3)$$

We adopt the L1 Chamfer Distance (L1-CD) (Fan, Su, and Guibas 2017) (refer to **Eq. (5)**) to measure the error between the sub-decoder's prediction and the actual features. For example, at layer $l$ during time $t$, the error is denoted as $\mathcal{L}_{CD}\left(\mathbf{E}_l^{(t)}, \overleftarrow{\mathbf{D}}_l^{(t)}\right)$. Subsequently, the error correction signal $\overleftarrow{\mathbf{G}}_l^{(t-1)}$ is quantified by computing the gradient of the L1-CD loss with respect to the previous feature $\mathbf{E}_l^{(t-1)}$:

$$\mathbf{G}_l^{(t-1)} = \nabla_{\mathbf{E}_l^{(t-1)}} \mathcal{L}_{CD}\left(\mathbf{E}_{l-1}^{(t-1)}, \mathbf{D}_{l-1}^{(t-1)}\right). \quad (4)$$

Such a feedback error correction is grounded in a theoretical hypothesis: the multi-level (multi-frequency) natural signals that are ubiquitous in the environment interact with the point cloud through hidden physical causes (e.g., altering its shape and color, introducing noise, causing missing points). The brain-inspired encoder minimizes the prediction error $\mathcal{L}_{CD}$ of layers, causing high-level features $\mathbf{E}_l^{(t)}$ to force shallow features for correcting themselves toward a state of "better predicted by high-levels". After learning over a long enough period, the updated parameters become capable of predicting these hidden physical principles at each level, thereby reconstructing the true object point clouds. After Eq. (4), the error correction $\overleftarrow{\mathbf{G}}_l^{(t-1)}$ is controlled by the parameter $\alpha$ to adjust the strength of error gradients.

## Training details and losses

The network uses the L1-CD as its loss function, as shown in **Eq. (5)**:

$$\mathcal{L}_{CD}(\mathcal{P}, \mathcal{G}) = \frac{1}{|\mathcal{P}|}\sum_{\mathbf{x}\in\mathcal{P}} \min_{\mathbf{y}\in\mathcal{G}} \|\mathbf{x}-\mathbf{y}\|_1^2 + \frac{1}{|\mathcal{G}|}\sum_{\mathbf{y}\in\mathcal{G}} \min_{\mathbf{x}\in\mathcal{P}} \|\mathbf{y}-\mathbf{x}\|_1^2, \quad (5)$$

where $\|\cdot\|_1^2$ represents the L1-distance, $\mathcal{P}_{pred}$ and $\mathcal{G}_{gt}$ are the predicted and Ground Truth (GT) point sets, respectively.

The feedback guidance received by each layer in BRICE are analogous to the features generated by deeper brain regions, which is similar to "drawing upon prior experience". Therefore, to better bridge the gap with the brain's prior knowledge, in the experiments of this paper we conduct pre-training on BRICE. We first connect BRICE to a simple classification head and perform pre-training for the point cloud classification task on LHD 3D dataset (Chen et al. 2023a). The pre-trained BRICE module is then transplanted into existing point cloud completion networks, followed by training and inference for the 3D completion task. During both the classification pre-training and the afterward training, we sum the reconstruction errors across all recurrent time steps $\mathrm{T}$ to form the encoder's internal optimization objective as $\mathcal{L}_{BRICE}$:

$$\mathcal{L}_{BRICE} = \sum_{l=1}^{3}\sum_{t=1}^{T} \mathcal{L}_{CD}\left(\mathbf{E}_l^{(t)}, \overleftarrow{\mathbf{D}}_l^{(t)}\right). \quad (6)$$

In all experiments in this paper, we set $\mathrm{T}=3$. When inserting BRICE into a completion network, we directly descend the original decoder. The loss $\mathcal{L}_{Decoder}$ in the decoder is generally calculated as the L1-CD difference between the completed point cloud $\mathbf{E}_{out}$ and the corresponding GT $\mathbf{E}_{gt}$:

$$\mathcal{L}_{Decoder} = \mathcal{L}_{CD}(\mathbf{E}_{out}, \mathbf{E}_{gt}). \quad (7)$$

The overall loss is calculated as $\mathcal{L}_{Total} = \mathcal{L}_{BRICE} + \mathcal{L}_{Decoder}$, which is the sum of two sub-losses.

# Experiments

## Datasets and training details

Four datasets were used in experiments, including The Labeled Hybrid Dataset (LHD) (Chen et al. 2023a), MVP2K (Pan et al. 2021), PCN (Yuan et al. 2018), and PlantPCom (Li et al. 2025). LHD is used for pre-training of BRICE under the classification task. LHD contains 235,175 point clouds (99.8% are man-made objects) from 87 categories. We evaluated classification performance using the Top-1 Accuracy metric (Deng et al. 2009). The MVP2K dataset contains 100,000 point clouds from 16 man-made target categories. Following the conventions of existing works that adopted MVP2K, we used L2 Chamfer Distance (L2-CD) to evaluate the completion performance. The PCN dataset is a point cloud completion benchmark built on ShapeNet dataset, comprising 30,974 point clouds from 8 man-made categories. Following the convention, we used L1-CD to assess the completion performance on PCN. To validate BRICE's generalization performance on natural objects, we also conducted experiments on the PlantPCom crop point cloud dataset, which contains 20,800 plant point clouds from 5 species. We used L1-CD to evaluate results on PlantPCom. Detailed description of the datasets is in **Appendix A**.

All experiments in this paper are conducted using PyTorch 2.4.1 on a workstation equipped with dual AMD EPYC 7513 CPUs, 128GB of RAM, and four paralleled NVIDIA 4090 GPUs. All evaluated baselines/networks were trained and tested on the single GPU. The configuration of hyperparameters and training are detailed in **Appendix B**. The detailed efficiency statistics including the speeds of inference, the FLOPs in inference, the parameters scales are referred to **Appendix C**.

## Results

**Pre-training for classification (LHD Dataset).** Initially, we pre-trained the BRICE encoder on the LHD dataset for the point cloud classification task. A simple fully-connected head is appended after BRICE to form a classification network. To prove that our BRICE can learn 3D semantics, we compared this simple network with several popular models on the testing set of LHD in **Table 1**. BRICE achieves the best Top-1 Accuracy at 78.11%, outperforms the second best—CRA-PCN baseline by 0.7%.

**Evaluation on MVP2K Dataset.** On this dataset, we integrated BRICE into seven mainstream completion baselines, including SnowflakeNet (Xiang et al. 2021), SeedFormer (Zhou et al. 2022), AnchorFormer (Chen et al. 2023b), PointAttN (Wang et al. 2024), ODGNet (Cai et al. 2024), CRA-PCN (Rong et al. 2024), and SymmCompletion (Yan et al. 2025) to see whether BRICE can improve point cloud completion. How we seamlessly replace their original encoder with BRICE for each baseline are shown in **Appendix D**. On each baseline, two different training modes were implemented. The first mode (called "+BRICE") directly inserts untrained BRICE into the baseline and trains the total network from scratch on the point cloud completion task. The second mode (called "+BRICE+Pre") transplants pre-trained BRICE (on LHD) into the baselines and fine-tunes the total networks on the completion task. **Table 2** presents the results of our comparative experiments, wherein the BRICE is embedded into 7 representative baselines. On the average L2-CD, BRICE enhances point cloud completion performance of all seven baselines on the MVP2K dataset. In addition, pre-training on the classification further improves BRICE across all seven baselines. For example, after integration of pre-trained BRICE, the L2-CD of PointAttN model decreases from 6.95 to 5.55 ($\times 10^{-4}$). The full quantitative results on all categories of MVP2K dataset can be found in **Appendix E**. Some representative qualitative results are visualized in **Fig. 3**, in which the BRICE-integrated baseline usually outperforms the original baseline at local areas that have structural discontinuities (e.g., the edges of the guitar, and the slender legs of the chair).

| **Baselines** | **PointNet++** | **PointAttN** | **CRA-PCN** | **Symm-Completion** | **BRICE** |
|---|---|---|---|---|---|
| **Top1 Acc ↑** | 75.56 | 68.97 | 77.41 | 75.15 | **78.11** |

Table 1: Quantitative comparison for point cloud classification between BRICE and several popular methods on the LHD dataset. Best is in bold; second-best underlined.

| **Baselines** | **Representative categories** | | | | | **Avg. ↓** |
|---|---|---|---|---|---|---|
| | **Air-plane** | **Lamp** | **Sofa** | **Table** | **Water-craft** | |
| SnowflakeNet | 2.11 | **7.43** | 7.51 | **7.33** | 5.12 | 6.05 |
| +BRICE | **2.10** | 7.49 | 7.32 | 7.49 | **4.97** | 6.05 |
| +BRICE+Pre | 2.13 | 7.57 | **7.16** | 7.37 | 5.06 | **6.04** |
| SeedFormer | 1.66 | 5.58 | 7.27 | 6.49 | 4.45 | 5.41 |
| +BRICE | 1.65 | 5.41 | 6.99 | **6.30** | 4.48 | 5.28 |
| +BRICE+Pre | **1.61** | **5.33** | **6.96** | 6.48 | **4.26** | **5.27** |
| AnchorFormer | 2.27 | 9.88 | 7.60 | 7.54 | 5.50 | 6.48 |
| +BRICE | **2.08** | 9.04 | 7.55 | 7.60 | 4.97 | 6.21 |
| +BRICE+Pre | 2.18 | **8.78** | **7.22** | **7.44** | **5.04** | **6.18** |
| PointAttN | 2.45 | 10.01 | 8.35 | 8.57 | 5.60 | 6.95 |
| +BRICE | **1.77** | 5.60 | 7.65 | **6.54** | 4.60 | 5.62 |
| +BRICE+Pre | 1.79 | **5.56** | **7.60** | 6.56 | **4.45** | **5.55** |
| ODGNet | 2.12 | 6.83 | 7.48 | 7.06 | 4.78 | 5.88 |
| +BRICE | 1.67 | 5.03 | **6.92** | 6.36 | 4.31 | 5.27 |
| +BRICE+Pre | **1.65** | **4.89** | 6.98 | **6.26** | **4.22** | **5.23** |
| CRA-PCN | 1.73 | 6.04 | 7.14 | 6.66 | 4.57 | 5.46 |
| +BRICE | 1.67 | 5.25 | 6.97 | 6.46 | 4.44 | 5.35 |
| +BRICE+Pre | **1.66** | **5.16** | **6.90** | **6.42** | **4.42** | **5.29** |
| SymmCompletion | 1.67 | 5.78 | 6.93 | 6.31 | 4.59 | 5.34 |
| +BRICE | **1.64** | 4.91 | 6.81 | **6.02** | 4.28 | 5.07 |
| +BRICE+Pre | 1.71 | **4.77** | **6.79** | 6.05 | **4.36** | **5.05** |

Table 2: Quantitative comparison on the MVP2K dataset by L2-CD (the lower the better). Omitted $\times 10^{-4}$ on all values for readability. Best in bold; second-best underlined. "+BRICE" denotes baselines integrated with BRICE, and "+BRICE+Pre" means integration of a pre-trained BRICE.

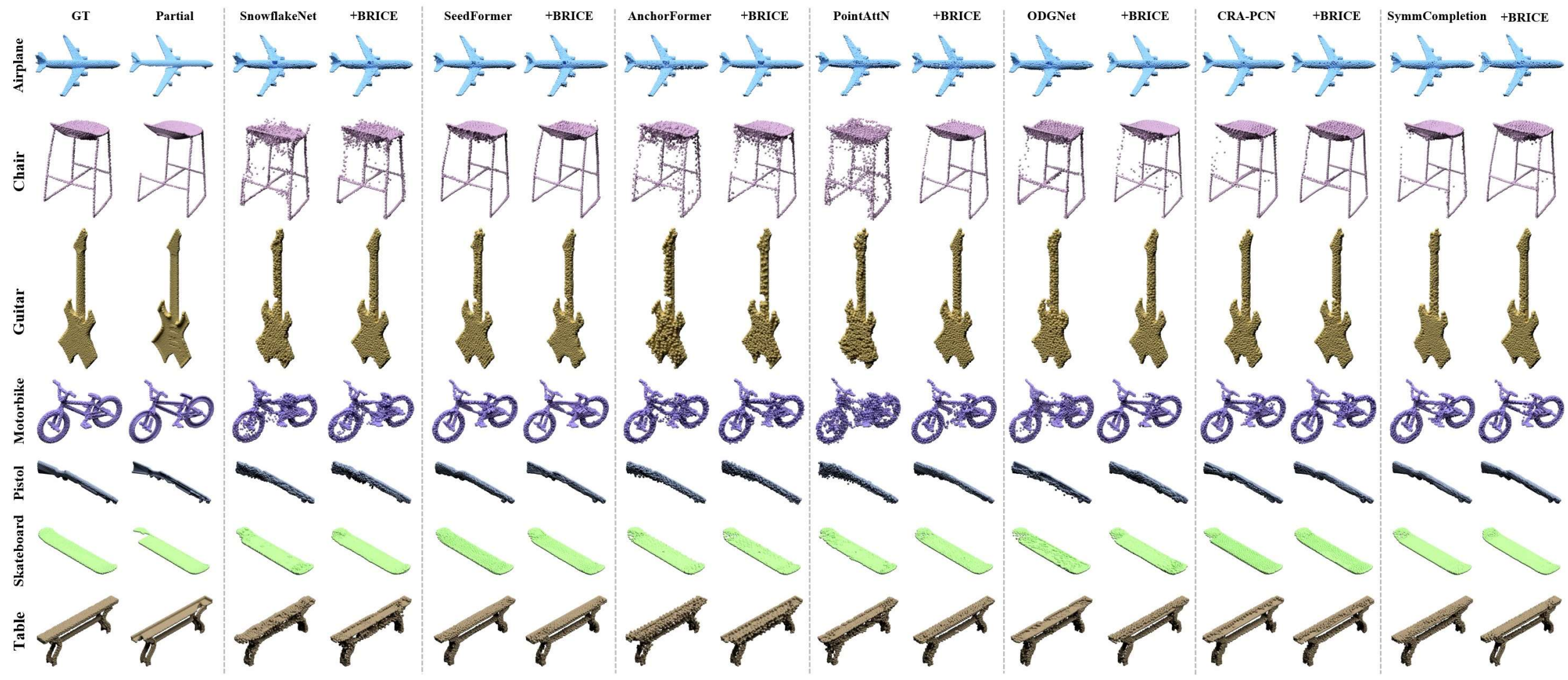

Figure 3: Qualitative results of the MVP2K Dataset. "GT" is the Ground Truth, "Partial" means the incomplete point cloud input. For saving space, "+BRICE" means the best result across the baseline with BRICE and with pretrained BRICE.

| Baselines | Categories | | | | | Avg. ↓ |
|---|---|---|---|---|---|---|
| | Cabinet | Chair | Lamp | Couch | Table | |
| SnowflakeNet | 9.46 | 8.07 | **6.27** | 9.55 | 6.65 | 7.40 |
| +BRICE | 9.38 | 7.92 | 6.28 | 9.31 | **6.60** | 7.30 |
| +BRICE+Pre | **9.37** | **7.68** | 6.32 | **8.99** | 6.64 | **7.25** |
| SeedFormer | 8.85 | 6.95 | 5.32 | 8.55 | 5.87 | 6.61 |
| +BRICE | 8.79 | 6.80 | 5.25 | 8.61 | 5.81 | 6.53 |
| +BRICE+Pre | **8.73** | **6.75** | **5.08** | **8.50** | **5.79** | **6.46** |
| AnchorFormer | 9.10 | 7.28 | 5.56 | 8.66 | 6.17 | 6.79 |
| +BRICE | **9.01** | 7.28 | 5.62 | 8.64 | 6.11 | 6.76 |
| +BRICE+Pre | 9.10 | **7.10** | **5.54** | **8.63** | **6.06** | **6.72** |
| PointAttN | 9.15 | 7.41 | **5.48** | 9.30 | 6.29 | 6.93 |
| +BRICE | 9.08 | 7.38 | 5.73 | 8.91 | **6.25** | 6.87 |
| +BRICE+Pre | **8.99** | **7.29** | 5.82 | **8.78** | 6.35 | **6.83** |
| ODGNet | 8.73 | **6.82** | **5.18** | **8.44** | **5.81** | **6.49** |
| +BRICE | 8.80 | 7.48 | 5.53 | 8.92 | 6.12 | 6.81 |
| +BRICE+Pre | **8.72** | 7.23 | 5.62 | 8.74 | 6.02 | 6.72 |
| CRA-PCN | 8.77 | 6.92 | 5.46 | 8.59 | 6.27 | 6.56 |
| +BRICE | **8.69** | 6.74 | 5.15 | 8.44 | 5.75 | 6.46 |
| +BRICE+Pre | 8.73 | **6.73** | **4.99** | **8.33** | **5.74** | **6.42** |
| SymmCompletion | **8.52** | **6.59** | 5.04 | **8.30** | **5.72** | **6.33** |
| +BRICE | 8.70 | 6.83 | **4.97** | 8.50 | 5.87 | 6.48 |
| +BRICE+Pre | 8.70 | 6.83 | **4.97** | 8.43 | 5.81 | 6.45 |

Table 3: Quantitative comparison on the PCN dataset by L1-CD (the lower the better). All values are omitted by $\times 10^{-3}$ for readability. Best in bold, second-best underlined.

**Evaluation on PCN Dataset.** On the PCN dataset, we integrated BRICE into the same seven completion baselines evaluated on the MVP2K dataset. As shown in **Table 3**, among all seven baselines, BRICE significantly improves the L1-CD performance of five models (SnowflakeNet, SeedFormer, AnchorFormer, PointAttN, and CRA-PCN); only ODGNet and SymmCompletion show a decline after integration. Visualizations of some qualitative examples can be found at **Fig. A2** in **Appendix F**, showing that BRICE improves completion on several objects with smoother surfaces and more realistic point distributions.

| Baselines | Categories | | | | | Avg.↓ |
|---|---|---|---|---|---|---|
| | Tomato | Tobacco | Sorghum | Ara-bidopsis | Maize | |
| SnowflakeNet | 4.11 | **4.53** | **3.15** | **3.87** | 9.10 | 5.45 |
| +BRICE | **4.04** | 4.84 | 3.26 | 4.28 | **8.08** | **5.26** |
| SeedFormer | 4.34 | 5.14 | 2.90 | 4.34 | 8.03 | 5.33 |
| +BRICE | **3.61** | **4.19** | **2.48** | **3.78** | **6.16** | **4.31** |
| AnchorFormer | 5.06 | 6.21 | 3.51 | 6.02 | 11.66 | 7.09 |
| +BRICE | **4.89** | **6.02** | **3.35** | **5.93** | **11.55** | **6.95** |
| PointAttN | 6.21 | 7.13 | 4.06 | 5.31 | 24.45 | 11.32 |
| +BRICE | **6.05** | **6.73** | **3.49** | **5.02** | **18.60** | **9.36** |
| ODGNet | 4.36 | 5.12 | 3.35 | 4.30 | 7.09 | 5.11 |
| +BRICE | **3.64** | **4.17** | **2.49** | **3.64** | **5.49** | **4.09** |
| CRA-PCN | 3.71 | 4.27 | 2.55 | 3.88 | 6.40 | 4.44 |
| +BRICE | **3.50** | **4.02** | **2.43** | **3.76** | **5.82** | **4.14** |
| SymmCompletion | 3.69 | 4.29 | 2.51 | 3.81 | 6.74 | 4.52 |
| +BRICE | **3.59** | **4.14** | **2.36** | **3.61** | **6.13** | **4.24** |

Table 4: Quantitative comparison results of PlantPCom dataset by L1-CD. Omitted by $\times 10^{-3}$ for readability.

On the PCN dataset, the undesirable adaptation of BRICE on ODGNet may relate to the dictionary-guided decoder of ODGNet that strongly abstracts common morphological features from all objects of the same category as a strong prior knowledge to facilitate 3D completion. This decoder design is well-suited for man-made objects (such as planes, vehicles, etc.) from which shared structural knowledge can be easily generalized, and the limited diversity of objects (only 8 classes) of PCN further amplifies the effect of the decoder. Adding BRICE to ODGNet probably breaks the formation of shared knowledge. BRICE's failing to improve SymmCompletion may due to its tailored decoder for 3D geometric symmetry as prior knowledge. Given that all 8 object classes in the PCN dataset are artificial objects with strong bilateral symmetry, the integration of BRICE disrupts the symmetry prior, leading to a decline in performance.

**Evaluation on PlantPCom Dataset.** To evaluate BRICE's performance on irregular natural objects, we also tested the point cloud completion models with BRICE on crop dataset—PlantPCom. Since pre-training on the LHD dataset introduces prior knowledge of symmetrical man-made objects into the encoder, and the knowledge contradicts with natural plants, we use BRICE without pre-training for PlantPCom. The quantitative results across seven evaluated point cloud completion baselines on PlantPCom are shown in **Table 4**, in which all models achieve better average L1-CD after inserting BRICE. In contrast to the performance on PCN data, ODGNet (+BRICE) not only becomes the best performer among all models but also has a significant improvement by $1.02\times10^{-3}$ in L1-CD using BRICE. Qualitative visualizations can be found at **Fig. A3** in **Appendix G**, showing that BRICE's effectiveness on help recovering finer details of plant organs.

The quantitative and qualitative results of multiple datasets show that our BRICE is general enough to further enhance mainstream completion networks on 3D point clouds of both man-made and natural objects.

## Ablation Studies

We conduct ablation study of BRICE using CRA-PCN as the baseline model on the MVP2K dataset (the Full model is "CRA-PCN+BRICE+Pre"), and the comparative results of different variants of the BRICE are shown in **Table 5**.

**Effectiveness of Feedback.** The main feedback pathway of BRICE encoder simulates the guidance signals from deeper brain regions to the shallow V1. When the feedback pathway is removed (i.e., $\lambda = 0$), this BRICE variant (w/o Main feedback) degenerates into a feedforward model, and its average L2-CD becomes merely on par with the original baseline (Plain model). This indicates that the supervision from deep features significantly augments the network's ability to infer missing regions in 3D point clouds.

**Effectiveness of Reconstruction errors.** To verify the usefulness of another auxiliary signal in the feedback pathway—the reconstruction errors in PC, we ablate the component that performs dynamic gradient correction using reconstruction errors (i.e., $\alpha = 0$). Under this configuration (w/o Recons), the model's average L2-CD error deteriorates to $5.48\times10^{-4}$, which is even worse than the plain model. This proves that the PC theory serves as a positive driving force for the dynamic updates of BRICE.

| **Variants** | Plain model | w/o Main feedback | w/o Recons | w/o Pretraining | Full model |
|---|---|---|---|---|---|
| **Avg. ↓** | 5.46 | 5.46 | 5.48 | <u>5.35</u> | **5.29** |

Table 5: Ablation experiments for CRA-PCN+BRICE on the MVP2K by L2-CD. Omitted by $\times10^{-4}$ for readability.

| **T** | 0 | 1 | 2 | 3 | 4 | 5 |
|---|---|---|---|---|---|---|
| **Avg. L2-CD ↓** | 5.46 | 5.39 | 5.36 | **5.29** | <u>5.34</u> | 5.35 |

Table 6: The tuning experiment of the recurrent time for CRA-PCN+BRICE on the MVP2K. Omitted by $\times10^{-4}$ for readability.

**Effectiveness of Pretraining.** Initializing network weights with cross-task knowledge has a positive effect on improving BRICE's completion performance on the same type of objects. As shown in **Table 5**, the full pretrained BRICE performs better than the variant trained from scratch (w/o Pretraining).

**The recurrent time T.** On the "CRA-PCN+BRICE+Pre" full model, we also have tuned the value of recurrent time T for the best performance. According to the tuning result in **Table 6**, with the increasing recurrent time, the performance first goes up and then falls, reaching optimal at 3. This result resembles the phenomenon when the brain generates ICs; i.e., visual illusions begin with recurrent loops but tend to disappear after a certain period of time. For humans, the long persistence of ICs is also unrealistic due to constant environmental changes or continuous saccadic eye movements, the completion derived from ICs often disappears after a few seconds (Ramachandran et al. 1994). To ensure BRICE having the optimal performance, the recurrent time is set to 3 throughout the experiments in this paper.

# Conclusion

Inspired from the IC mechanism in the V1L2/3 region of brain, we propose BRICE—a plug-and-play encoder for enhancing point cloud completion. BRICE can be seamlessly inserted into (replacing the original encoder) various mainstream point cloud completion networks to improve performance. BRICE is currently only well suited for baselines with clear encoder-decoder architectures. The applicability of BRICE on the networks that either lack clear hierarchy or have irregularly stacked layers are still unexplored. In future work, we will try reducing the parameters of BRICE and further testing it on more diversified 3D datasets and designing new versions for unified vision tasks.

# Appendix to BRICE: Brain-inspired Recurrent Illusory Contour Encoder for Point Cloud Completion

Our code: https://github.com/wangyunlei-wyl/BRICE.

## A. The Details of the datasets

**LHD Point Cloud Classification Dataset** (Chen et al. 2023a)**.** The Labeled Hybrid Dataset (LHD) is a point cloud classification benchmark; we used it to pre-train BRICE on the classification task. LHD contains 235,175 point clouds (99.8% are man-made objects) from 87 categories, in which 183,704 are used for training and the remaining 51,471 for validation. Input point clouds were uniformly resized to 2,048 points for training. We evaluated classification performance using the Top-1 Accuracy metric (Deng et al. 2009) on LHD.

**MVP2K Point Cloud Completion Dataset** (Pan et al. 2021)**.** The MVP2K dataset contains 100,000 point clouds from 16 target semantic categories, in which 62,400 point clouds are used for training and the rest are used for testing. Each object is rendered from 26 camera viewpoints uniformly distributed on a sphere; the parts of the point cloud not visible from a given viewpoint are directly removed, resulting in 26 incomplete point clouds with different poses. Additionally, both each incomplete point cloud and the corresponding complete ground-truth point cloud were fixed to 2,048 points. Following the comparison conventions of most existing works, we used L2 Chamfer Distance (L2-CD) to evaluate the completion performance.

**PCN Point Cloud Completion Dataset** (Yuan et al. 2018)**.** The PCN dataset is a point cloud completion benchmark built on ShapeNet dataset, comprising 30,974 point clouds from 8 man-made categories. Among these, 28,974 point clouds are used for training, 800 for validation, and the remaining 1,200 for testing. All input point clouds were fixed at 2,048 points. Following the evaluation conventions of most existing studies, we used L1-CD to assess point cloud completion performance on this dataset.

**PlantPCom Plant Point Cloud Completion Dataset** (Li et al. 2025)**.** To validate BRICE's generalization performance on natural 3D objects, we conducted experiments on the PlantPCom crop point cloud dataset. PlantPCom contains a total of 20,800 3D plants from 5 species (tomato, tobacco, Arabidopsis, sorghum, and maize). The dataset is split into training, validation, and testing sets with an 8:1:1 ratio, with each input point cloud fixed at 2,048 points. Similar to MVP2K, the incomplete point clouds of PlantPCom are generated from high-precision 3D plant models across 20 different viewpoints. For this dataset, we used L1-CD to evaluate point cloud completion performance.

| Baselines | Params (M) | Flops (G) | Inference time (ms) | Training time (h) |
|---|---|---|---|---|
| CRA-PCN | 5.68 | 4.63 | 39.00 | 13.6 |
| +BRICE | 17.51 | 20.03 | 40.58 | 19.2 |
| PointAttN | 19.00 | 6.17 | 12.46 | 8.4 |
| +BRICE | 21.00 | 22.09 | 35.06 | 16.2 |
| ODGNet | 4.48 | 4.71 | 15.35 | 7.8 |
| +BRICE | 23.92 | 22.52 | 44.19 | 25.6 |
| SnowflakeNet | 18.79 | 6.30 | 9.13 | 5.4 |
| +BRICE | 31.80 | 22.29 | 26.49 | 12.8 |
| AnchorFormer | 30.46 | 6.60 | 46.46 | 21.0 |
| +BRICE | 44.23 | 23.13 | 74.14 | 19.9 |
| SeedFormer | 2.64 | 5.86 | 20.77 | 11.0 |
| +BRICE | 17.54 | 23.50 | 32.88 | 19.8 |
| SymmCompletion | 6.64 | 7.97 | 17.82 | 23.3 |
| +BRICE | 19.36 | 24.67 | 40.11 | 30.2 |

Table A1: Detailed efficiency statistics about the evaluated baselines on the MVP2K dataset.

## B. The training details

To ensure fairness, for all experiments in this paper unless otherwise specified, the three parameter values in Eq. (1) were set as $\alpha = 0.01$, $\beta = 0.8$, and $\lambda = 0.1$. For pre-training of BRICE, we adopted the same configuration as suggested in PointGPT (Chen et al. 2023a).

For all baselines running on both MVP2K and PlantPCom datasets, we used the Adam optimizer (Diederik 2014) with $\boldsymbol{\beta}_1 = 0.9$ and $\boldsymbol{\beta}_2 = 0.999$, and weight decay at 0.001. The learning rate is 0.0001 for MVP2K and 0.001 for PlantPCom. The training batch size is fixed at 48 for MVP2K, and 32 for PlantPCom. The learning rate warmup mechanism was applied to ensure stable model convergence. On MVP2K and PlantPCom, all models were trained for 100 epochs, and the models having the lowest training loss were used for testing.

For the PCN dataset, all baselines were trained with the recommended hyperparameter settings in their publications (Xiang et al. 2021; Zhou et al. 2022; Chen et al. 2023b; Wang et al. 2024; Cai et al. 2024; Rong et al. 2024; Yan et al. 2025), respectively.

| Baselines | Categories | | | | | | | | | | | | | | | | Avg. ↓ |
|---|---|---|---|---|---|---|---|---|---|---|---|---|---|---|---|---|---|
| | Air | Cab | Car | Cha | Lam | Sof | Tab | Wat | Bed | Ben | Boo | Bus | Gui | Mot | Pis | Ska | |
| SnowflakeNet | 2.11 | 7.88 | 5.94 | 7.53 | **7.43** | 7.51 | **7.33** | 5.12 | 11.32 | 4.70 | **8.13** | 4.52 | 1.21 | **4.05** | **3.31** | **3.77** | 6.05 |
| +BRICE | **2.10** | **7.87** | 5.86 | 7.50 | 7.49 | 7.32 | 7.49 | **4.97** | 11.51 | 4.82 | 8.23 | **4.48** | **1.20** | 4.10 | 3.58 | 4.03 | 6.05 |
| +BRICE+Pre | 2.13 | 7.94 | **5.85** | **7.40** | 7.57 | **7.16** | 7.37 | 5.06 | **10.83** | **4.67** | 8.32 | 4.62 | 1.26 | 4.21 | 3.44 | 4.36 | **6.04** |
| SeedFormer | 1.66 | 7.68 | 5.73 | 6.70 | 5.58 | 7.27 | 6.49 | 4.45 | **10.48** | 4.59 | 7.29 | 4.37 | 1.04 | 3.53 | **2.96** | 2.07 | 5.41 |
| +BRICE | 1.65 | **7.60** | 5.56 | **6.20** | 5.41 | 6.99 | **6.30** | 4.48 | 10.89 | **4.40** | 7.11 | **4.20** | **1.01** | **3.47** | 3.07 | 2.10 | 5.28 |
| +BRICE+Pre | **1.61** | 7.67 | **5.53** | 6.35 | **5.33** | **6.96** | 6.48 | **4.26** | 10.56 | 4.42 | **7.01** | 4.25 | 1.06 | 3.53 | 3.19 | **1.88** | **5.27** |
| AnchorFormer | 2.27 | 7.85 | 5.71 | 7.94 | 9.88 | 7.60 | 7.54 | 5.50 | 13.51 | 5.29 | 9.00 | 4.47 | 1.23 | 4.03 | 3.78 | 2.91 | 6.48 |
| +BRICE | **2.08** | **7.75** | 5.65 | 7.57 | 9.04 | 7.55 | 7.60 | **4.97** | 12.70 | 5.04 | **8.60** | **4.39** | 1.27 | 3.98 | **3.21** | 2.87 | 6.21 |
| +BRICE+Pre | 2.18 | 8.54 | **5.54** | **7.30** | **8.78** | **7.22** | **7.44** | 5.04 | **12.11** | **4.98** | 8.79 | 4.51 | **1.21** | **3.97** | 3.33 | **2.83** | **6.18** |
| PointAttN | 2.45 | 8.81 | 6.15 | 8.43 | 10.01 | 8.35 | 8.57 | 5.60 | 12.77 | 5.46 | 10.41 | 4.92 | 1.49 | 4.47 | 4.11 | 3.59 | 6.95 |
| +BRICE | **1.77** | **8.08** | 6.10 | 6.72 | 5.60 | 7.65 | **6.54** | 4.60 | 11.03 | 4.67 | 7.91 | 4.67 | 1.17 | 3.79 | 3.29 | 2.15 | 5.62 |
| +BRICE+Pre | 1.79 | 8.18 | **6.05** | **6.61** | **5.56** | **7.60** | 6.56 | **4.45** | **10.56** | **4.46** | **7.58** | **4.60** | **1.16** | **3.67** | **2.93** | **2.14** | **5.55** |
| ODGNet | 2.12 | **7.84** | 5.95 | 7.40 | 6.83 | 7.48 | 7.06 | 4.78 | 11.89 | 4.71 | 7.63 | 4.62 | 1.18 | 4.11 | 3.42 | 2.25 | 5.88 |
| +BRICE | 1.67 | 7.91 | **5.92** | **6.43** | 5.03 | **6.92** | 6.36 | 4.31 | **9.29** | 4.06 | 7.25 | **4.50** | 1.10 | **3.59** | 3.10 | 2.21 | 5.27 |
| +BRICE+Pre | **1.65** | 7.90 | 5.93 | 6.44 | **4.89** | 6.98 | **6.26** | **4.22** | 9.54 | **3.86** | **6.79** | 4.55 | **1.05** | 3.60 | **2.92** | **2.11** | **5.23** |
| CRA-PCN | 1.73 | **7.70** | **5.70** | **6.40** | 6.04 | 7.14 | 6.66 | 4.57 | 10.44 | 4.65 | **7.02** | **4.37** | **1.01** | **3.48** | 3.00 | 2.19 | 5.46 |
| +BRICE | 1.67 | 7.77 | 5.80 | 6.55 | 5.25 | 6.97 | 6.46 | 4.44 | 9.77 | 4.34 | 7.22 | 4.46 | 1.10 | 3.61 | **2.92** | 2.13 | 5.35 |
| +BRICE+Pre | **1.66** | 7.88 | 5.77 | 6.47 | **5.16** | **6.90** | **6.42** | **4.42** | **9.60** | **4.13** | 7.32 | 4.45 | 1.09 | 3.58 | 3.02 | **2.03** | **5.29** |
| SymmCompletion | 1.67 | 7.70 | **5.54** | **6.03** | 5.78 | 6.93 | 6.31 | 4.59 | 11.29 | 4.63 | 7.12 | 4.37 | 1.09 | **3.47** | 2.97 | 2.15 | 5.34 |
| +BRICE | **1.64** | **7.42** | 5.62 | 6.08 | 4.91 | 6.81 | **6.02** | **4.28** | **9.34** | **3.79** | 6.66 | **4.29** | 0.96 | 3.49 | 2.84 | **2.12** | 5.06 |
| +BRICE+Pre | 1.71 | 7.47 | 5.59 | 6.05 | **4.77** | **6.79** | 6.05 | 4.36 | 9.60 | 3.86 | **6.42** | 4.40 | **0.95** | 3.48 | **2.67** | 2.14 | **5.05** |

Table A2: Detailed quantitative comparison results of L2-CD on the MVP2K dataset. All values are omitted by $\times 10^{-4}$ for readability. The best results are indicated in bold, and the second-best results are underlined. Here, “+BRICE” denotes baselines integrated with our BRICE encoder, and “+BRICE+Pre” indicates the utilization of a pre-trained BRICE encoder.

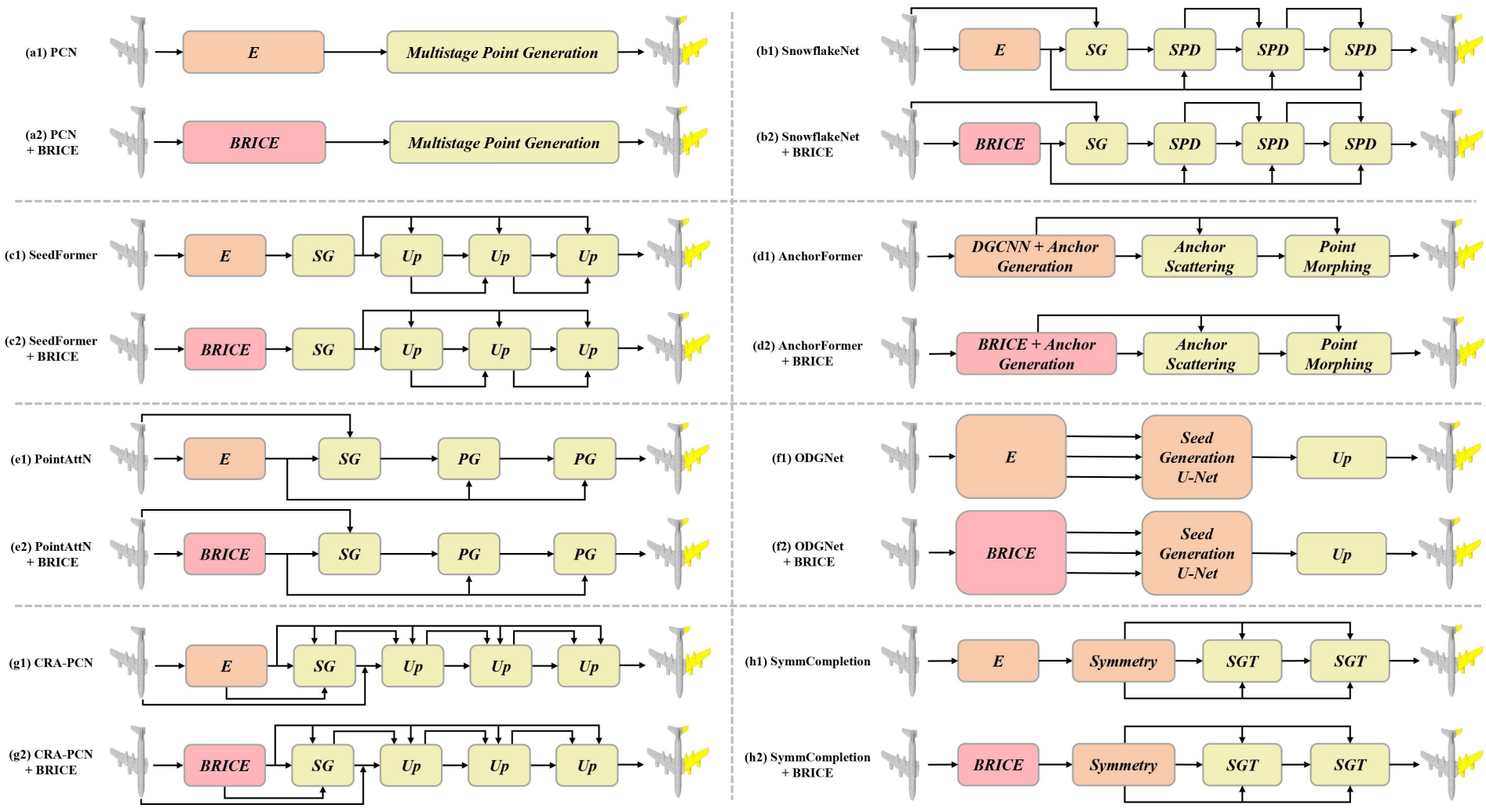

Figure A1: Workflow diagrams of 8 networks for point cloud completion, for each network we display both the original architecture and the BRICE-integrated architecture. “E” denotes the original encoder in the plain baseline. “SG” represents the Seed Generation module. “SPD” refers to the Snowflake Point Deconvolution module in SnowFlakeNet. “Up” stands for the Up-sampling layer. “PG” denotes the Point Generator module in the PointAttN network. SGT indicates the Symmetry-Guided Transformer module in the SymmCompletion network.

## C. Efficiency of learning

Take the MVP2K dataset for instance, the evaluated baselines (both with and w/o BRICE) took around 5.4 to 30.2 hours to finish training. The average inference time on a single point cloud for BRICE-enhanced baseline ranges from 9.13 ms to 74.14 ms. The parameters of BRICE are mainly due to the up-sampling and three transformer blocks. The full information about the parameters per baseline, the FLOPs in inference, and the training and inference speeds on MVP2K is collected in **Table A1**.

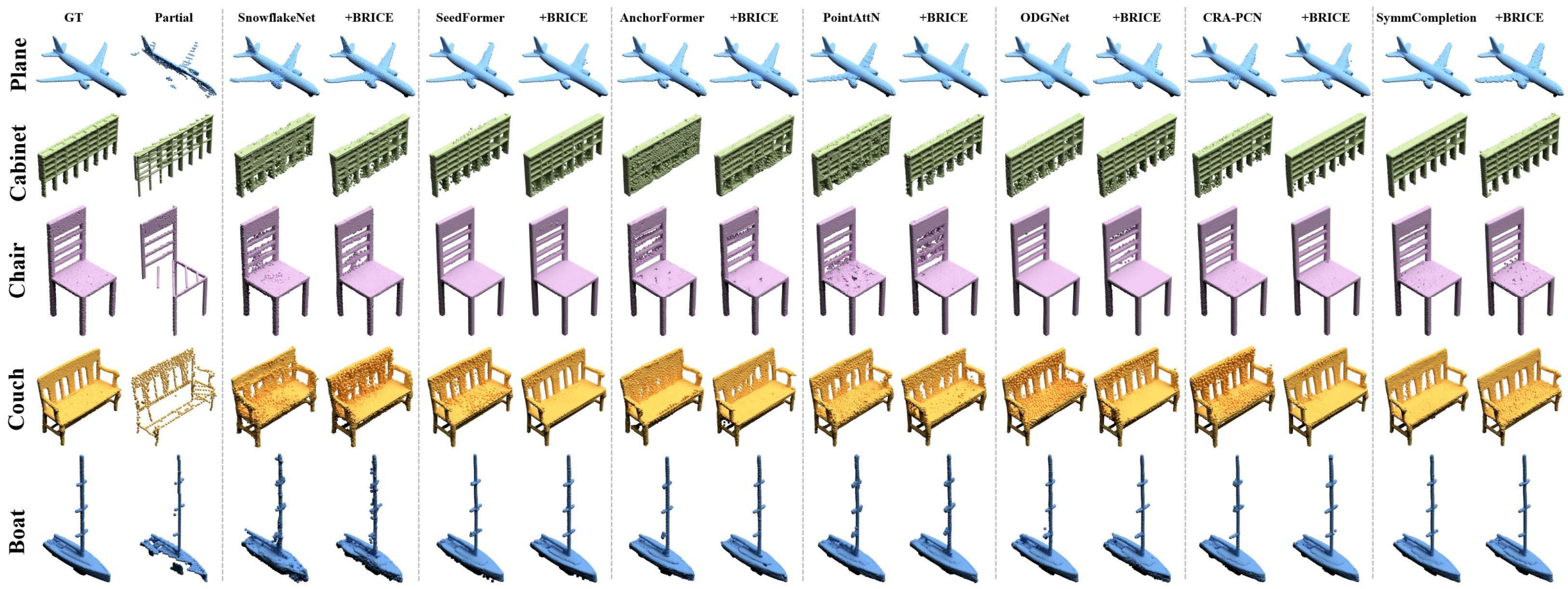

Figure A2: Qualitative results of the PCN Dataset. “GT” is the Ground Truth, “Partial” means the incomplete point cloud input. For saving space, “+BRICE” means the best result across the baseline with BRICE and with pretrained BRICE.

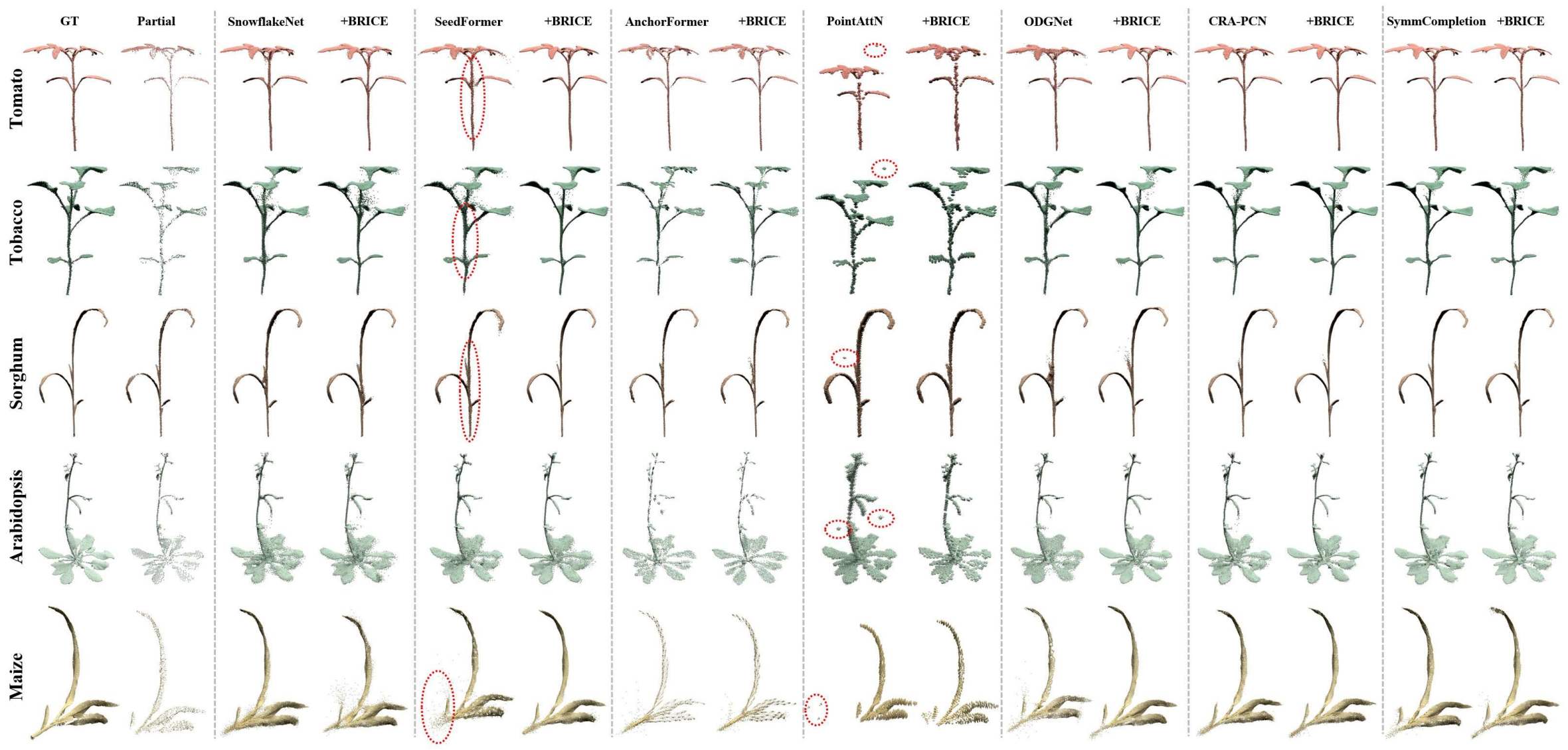

Figure A3: Qualitative results of the PlantPCom Dataset. “GT” is the Ground Truth, “Partial” means the incomplete point cloud input. “+BRICE” means integration of un-pretrained BRICE. The red dashed ellipses highlight the incorrectly recovered areas.

## D. Detailed architectures of evaluated baselines/networks

In this subsection, we visually demonstrate how we used BRICE as a plug-and-play module for enhancing the mainstream point cloud completion networks. **Fig. A1** compares the original architecture and the BRICE-integrated architecture on 8 mainstream point cloud completion baselines/networks (including PCN, SnowflakeNet, SeedFormer, AnchorFormer, PointAttN, ODGNet, CRA-PCN, and SymmCompletion), respectively.

## E. More quantitative results on MVP2K dataset

Due to space limitation in the manuscript, we present in **Table A2** the quantitative results of only five categories (Airplane, Lamp, Sofa, Table, and Watercraft) as well as the total average of the MVP2K dataset. To comprehensively demonstrate the effectiveness of the proposed BRICE encoder, we list the full quantitative results for all categories in **Table A2**.

## F. Qualitative results on PCN dataset

Visualizations of some qualitative examples of PCN dataset can be found at **Fig. A2**, showing that BRICE improves completion on several objects with smoother surfaces and more realistic point distributions. In **Fig. A2**, the plain SeedFormer and plain CRA-PCN exhibit structural gaps or rough surfaces when completing the top of "Couch" and the mast of "Boat"; however, after the help of BRICE, the two objects exhibit smoother surfaces and more realistic point distributions.

## G. Qualitative results on PlantPCom Dataset

**Fig. A3** visualizes some qualitative examples of the PlantPCom dataset. SeedFormer tends to produce a large amount of noise in the stem region (highlighted in red dashed ellipses), whereas SeedFormer+BRICE significantly optimizes the morphology of stems and suppresses noise. PointAttN creates several separated noise clusters on tomato, tobacco, and maize (red dashed ellipses); and the integration of BRICE cleans those undesirable clusters.